\documentclass{article}

\usepackage[accepted]{icml2023}
\usepackage{microtype}
\usepackage{graphicx}
\usepackage{subfigure}
\usepackage{booktabs} 

\usepackage{hyperref}

\usepackage{icml2023}

\usepackage{amsmath}
\usepackage{amssymb}
\usepackage{mathtools}
\usepackage{amsthm}

\usepackage[capitalize,noabbrev]{cleveref}

\theoremstyle{plain}

\theoremstyle{definition}

\theoremstyle{remark}

\usepackage[textsize=tiny]{todonotes}

\icmltitlerunning{Submission and Formatting Instructions for ICML 2023}

\begin{document}

\twocolumn[
\icmltitle{Inter-Annotator Agreement in the Wild: Uncovering Its Emerging Roles and Considerations in Real-World Scenarios}



\icmlsetsymbol{equal}{*}

\begin{icmlauthorlist}
\icmlauthor{NamHyeok Kim}{yyyc,equal}
\icmlauthor{Chanjun Park}{yyyc,equal}
\end{icmlauthorlist}
  
\icmlaffiliation{yyyc}{Upstage, Gyeonggi-do, Korea}


\icmlcorrespondingauthor{Chanjun Park}{chanjun.park@upstage.ai}

\icmlkeywords{Machine Learning, ICML}

\vskip 0.3in
]



\printAffiliationsAndNotice{\icmlEqualContribution} 

\begin{abstract}
Inter-Annotator Agreement (IAA) is commonly used as a measure of label consistency in natural language processing tasks. However, in real-world scenarios, IAA has various roles and implications beyond its traditional usage. In this paper, we not only consider IAA as a measure of consistency but also as a versatile tool that can be effectively utilized in practical applications. Moreover, we discuss various considerations and potential concerns when applying IAA and suggest strategies for effectively navigating these challenges.
\end{abstract}

\section{Introduction}
Inter-Annotator Agreement (IAA) is traditionally used in natural language processing (NLP) tasks as a measure of label consistency among annotators~\cite{artstein2017inter}. However, its role and implications extend beyond this customary usage. In this paper, we delve into the emerging functions of IAA in real-world scenarios. We propose that IAA not only serves as a consistency measure, but also as a tool for assessing the quality of annotators, predicting the complexity of documents, estimating model performance, and quantifying annotator abilities. Further, we address various considerations related to the use of IAA. These include the units of measurement, the relative nature of IAA scores, the influence of different annotator combinations and task characteristics, and the limitations of associating high IAA with superior data quality. Finally, we offer guidance on preparing for and managing situations with low IAA, providing valuable insights for those involved in data modeling.

\section{IAA in Real-World Scenarios}
Inter-Annotator Agreement (IAA) plays a critical role in assessing the quality of annotations and the reliability of annotated datasets in real-world applications.

\textbf{Firstly}, IAA serves as a criterion for selecting the most suitable group of annotators, thereby optimizing the data construction process~\cite{lingren2014evaluating}. For instance, from a pool of 100 annotators, selecting 10 with a high IAA score can enhance data consistency. Essentially, clustering annotators who exhibit similar behaviors using IAA can elevate data quality and decrease data cleansing expenses, leading to overall cost efficiency. \textbf{Secondly}, IAA aids in refining guidelines and improving annotator training~\cite{amidei2020identifying}. Regardless of the quality of data or the expertise of the annotators, generating consistently high-quality data can be challenging if the guidelines or training are inadequate. By scrutinizing the IAA score and its results, we can identify areas of confusion and subsequently enhance guidelines and training, leading to improved data quality. \textbf{Lastly}, IAA can act as a foundation for data cleansing~\cite{choi2023dmops}. Typically, achieving high-quality data in a single cycle is challenging, making data cleansing a frequent necessity. This process is often based on the model's performance. However, by employing IAA, data cleansing can be executed during the data construction phase itself, without relying on the model. For example, if specific areas consistently yield a low IAA score for a particular entity, we can modify those areas (via guideline revisions, enhanced training, or annotator improvement strategies) to construct data with higher consistency.

\section{Emerging Roles of IAA} 
In this section, we expand upon the emerging roles of Inter-Annotator Agreement (IAA) that extend beyond its traditional function of assessing label consistency. Specifically, we highlight innovative applications of IAA as a tool for evaluating annotator proficiency, predicting document complexity, estimating model accuracy, and measuring individual annotator capabilities.

\textbf{Firstly}, to assess annotator proficiency, we propose a method that involves analyzing individual IAA scores. The consistency in an annotator's labeling can serve as a valuable measure of their reliability and skill. This approach can help identify those who consistently produce high-quality annotations and those who may require additional training or support. \textbf{Secondly}, we suggest using IAA scores to predict the complexity of specific documents. A document that exhibits lower agreement levels among multiple annotators often signals greater annotation challenges or inherent ambiguity. Recognizing such documents allows for more effective resource allocation and can lead to improvements in overall model performance. \textbf{Thirdly}, IAA can be utilized to estimate model performance in comparison to human annotators. The degree of agreement between a model's predictions and human-annotated labels can provide a meaningful measure of the model's ability to understand and classify text accurately. \textbf{Fourthly}, IAA also facilitates the quantification of individual annotator abilities. By comparing IAA scores across different annotators, we can detect variations in expertise levels, enabling more tailored task assignments.

Through these proposed applications, IAA can provide invaluable insights into annotator proficiency, document complexity, model accuracy, and individual annotator capabilities. Thus, by effectively leveraging IAA, practitioners can significantly enhance the overall quality and efficiency of NLP annotation tasks.

\section{Considerations Regarding IAA} 
In this section, we delve into key considerations surrounding IAA, an essential component in data annotation processes.

Low IAA can arise from a variety of sources, such as complex tasks, low-quality raw data, limited understanding or skills among annotators, unsuitable pairings of annotators, ambiguous guidelines and training, incorrect annotation rules, or non-intuitive UI/UX of annotation tools. However, identifying the specific cause behind a low IAA can be intricate. Accurate identification of these causes allows us to systematically improve data quality during the data construction process, which typically involves a human-in-the-loop.

One potential approach to address this issue is to incorporate F1-scores as a metric. This could involve analyzing the data F1-score (which compares the predicted labels and the ground truth), and the model F1-score (which compares the model's predictions and the ground truth).

For example, if annotators A and B exhibit high IAA and their data F1-scores are also high, but annotaor C presents a low agreement and data F1-score, the issue might be attributed to annotaor C's proficiency. Conversely, if all annotaors A, B, and C have high IAA agreement, but the model performance is low, it may suggest that despite the annotators performing well, the model failed to learn the desired outcome, indicating a potential issue with the annotation rules. In another scenario, if the annotators generally demonstrate high agreement but exhibit low agreement for a specific task, we might question the complexity of that particular task.

When applying IAA, several additional considerations should be acknowledged. These include the requirement for all annotators to use the same data for IAA measurement, the fact that the degree of agreement can vary greatly depending on the unit of IAA measurement, and the limitation that IAA is best suited for tasks akin to classification. These considerations should be taken into account to effectively employ IAA.

\section{Dealing with Low IAA in the wild}
In this section, we provide strategic insights for preparing for and addressing situations of low IAA.

When low IAA arises due to the high difficulty of the task, the ideal solution usually lies in enhancing the annotators' experience. Given that altering the task itself is often unfeasible, it might be more advantageous to allow the annotators to accumulate experience over an extended period, and then undertake data cleansing, rather than resorting to immediate cleansing upon the emergence of low IAA~\cite{lommel2014assessing}.

If poor data quality results in low IAA, an obvious solution would be to eliminate low-quality data and replace it with high-quality data~\cite{rahm2000data}. However, this approach is often unworkable in practice. Given the difficulty associated with altering low-quality data, the feasible solutions usually involve refining annotation rules to increase their specificity or reinforcing training to elevate the level of agreement among annotators.

In situations where low IAA is attributed to annotators' inadequate understanding or skillset, the resolution should involve targeted education and feedback. Ambiguous guidelines and training can be rectified by revising the guidelines to eliminate potential ambiguity and enriching them with diverse examples.

When low IAA is a consequence of non-intuitive UI/UX of annotation tools, it is advisable to enhance the UI/UX to ensure annotators can intuitively comprehend and utilize the tool. If enhancing the UI/UX is not possible, detailed education about the annotation tool should be provided to ensure all annotators can use the tool consistently.

\section{Conclusions}
In this paper, we have explored the multifaceted roles and considerations of Inter-Annotator Agreement (IAA), moving beyond its conventional use as a measure of label consistency. Moving beyond positioning, we plan to conduct actual experiments and various validations in the future.


\nocite{langley00}

\bibliography{example_paper}

\begin{thebibliography}{6}
\providecommand{\natexlab}[1]{#1}
\providecommand{\url}[1]{\texttt{#1}}
\expandafter\ifx\csname urlstyle\endcsname\relax
  \providecommand{\doi}[1]{doi: #1}\else
  \providecommand{\doi}{doi: \begingroup \urlstyle{rm}\Url}\fi

\bibitem[Amidei et~al.(2020)Amidei, Piwek, and Willis]{amidei2020identifying}
Amidei, J., Piwek, P., and Willis, A.
\newblock Identifying annotator bias: A new irt-based method for bias
  identification.
\newblock 2020.

\bibitem[Artstein(2017)]{artstein2017inter}
Artstein, R.
\newblock Inter-annotator agreement.
\newblock \emph{Handbook of linguistic annotation}, pp.\  297--313, 2017.

\bibitem[Choi \& Park(2023)Choi and Park]{choi2023dmops}
Choi, E. and Park, C.
\newblock Dmops: Data management operation and recipes.
\newblock \emph{arXiv preprint arXiv:2301.01228}, 2023.

\bibitem[Lingren et~al.(2014)Lingren, Deleger, Molnar, Zhai, Meinzen-Derr,
  Kaiser, Stoutenborough, Li, and Solti]{lingren2014evaluating}
Lingren, T., Deleger, L., Molnar, K., Zhai, H., Meinzen-Derr, J., Kaiser, M.,
  Stoutenborough, L., Li, Q., and Solti, I.
\newblock Evaluating the impact of pre-annotation on annotation speed and
  potential bias: natural language processing gold standard development for
  clinical named entity recognition in clinical trial announcements.
\newblock \emph{Journal of the American Medical Informatics Association},
  21\penalty0 (3):\penalty0 406--413, 2014.

\bibitem[Lommel et~al.(2014)Lommel, Popovic, and
  Burchardt]{lommel2014assessing}
Lommel, A., Popovic, M., and Burchardt, A.
\newblock Assessing inter-annotator agreement for translation error annotation.
\newblock In \emph{MTE: Workshop on Automatic and Manual Metrics for
  Operational Translation Evaluation}, pp.\  31--37. Language Resources and
  Evaluation Conference Reykjavik, 2014.

\bibitem[Rahm et~al.(2000)Rahm, Do, et~al.]{rahm2000data}
Rahm, E., Do, H.~H., et~al.
\newblock Data cleaning: Problems and current approaches.
\newblock \emph{IEEE Data Eng. Bull.}, 23\penalty0 (4):\penalty0 3--13, 2000.

\end{thebibliography}
\bibliographystyle{icml2023}

\end{document}